\documentclass{svproc}
\usepackage[T5,T1]{fontenc}
\usepackage{url}
\usepackage{cite}
\usepackage{amsmath,amssymb,amsfonts}
\usepackage{algpseudocode}
\usepackage{algorithm}
\usepackage{algorithmicx}

\begin{document}
\mainmatter              % start of a contribution
\title{Evidence, Definitions and Algorithms regarding the Existence of Cohesive-Convergence Groups in Neural Network Optimization}
\titlerunning{Existence of Cohesive-Convergence Groups}  % abbreviated title (for running head)
\author{\fontencoding{T5}\selectfont Thi\d\ecircumflex{}n \Acircumflex{}n L. Nguy\~\ecircumflex{}n}
\authorrunning{\fontencoding{T5}\selectfont Thi\d\ecircumflex{}n \Acircumflex{}n L. Nguy\~\ecircumflex{}n} % abbreviated author list (for running head)
\tocauthor{\fontencoding{T5}\selectfont Thi\d\ecircumflex{}n \Acircumflex{}n L. Nguy\~\ecircumflex{}n}
\institute{
\email{thienannguyen.cv@gmail.com}}

\maketitle              % typeset the title of the contribution

\begin{abstract}
Understanding the convergence process of neural networks is one of the most complex and crucial issues in the field of machine learning. Despite the close association of notable successes in this domain with the convergence of artificial neural networks, this concept remains predominantly theoretical. In reality, due to the non-convex nature of the optimization problems that artificial neural networks tackle, very few trained networks actually achieve convergence. To expand recent research efforts on artificial-neural-network convergence, this paper will discuss a different approach based on observations of cohesive-convergence groups emerging during the optimization process of an artificial neural network. 
\end{abstract}
\section{Introduction}
This paper addresses three intuitive problems. First, when a neural network is trained on a dataset and starts converging around an optimal point, if the distance between two samples in this dataset is small enough, then the value of the corresponding objective function for these two samples either increases for both or decreases for both. Second, for classification problems, whether these two samples, separated by such a distance, contain information about the labels that this neural network is trying to predict. Third, whether set of these pairs of samples contains information about the underfitting or overfitting status of the neural network.

The neural network optimization is recognized as a concept with diverse implications in both structure and methodology. To mitigate the impact of this diversity on findings, the author focuses solely on definitions and concepts relevant to the observations present in this paper and introduces algorithms based on these definitions and concepts. Basically, the paper consists of three sections:
\begin{itemize}
\item Concepts and definitions. 
\item Experiments demonstrating the practicality of the concepts and definitions. 
\item Algorithms. 
\end{itemize}

\section{Definitions}
A convergence process of a neural network involves three components: a dataset $D=\{(x_0,y_0), .., (x_N, y_N)\}$, a neural network $F_{\theta}(\cdot): \mathbb{R}^{n} \rightarrow \mathbb{R}$ has $\theta$ as trainable parameters, and a stochastic training process $T^{k}(\cdot)$ return a trained neural network with $k$ representing the number of training steps. The training process $T^{k}$ satisfies the following condition:
\begin{itemize}
\item For any value of $\theta=\theta_0$, by which empirical risk of $F_{\theta_0}$ over $D_{train} \subsetneq D$ is equal to $c>0$ ($L(F_{\theta_0},D_{train})=c$), there exists $k_0$ such that $L(T^{k'}(F_{\theta_0}),D_{train}) < c, \forall k' > k_0$.
\end{itemize}

\subsection{Cohesive-Convergence Group}
Let $A_{d_0, d_1}$ denote the event $L(T^{K}(F_{\theta}),d_0) < L(T^{K+1}(F_{\theta}),d_0)$ and $L(T^{K}(F_{\theta}),d_1) < L(T^{K+1}(F_{\theta}),d_1)$ of a trial $(K,K+1)$. Similarly, let $B_{d_0, d_1}$ denote the event $L(T^{K}(F_{\theta}),d_0) > L(T^{K+1}(F_{\theta}),d_0)$ and $L(T^{K}(F_{\theta}),d_1) > L(T^{K+1}(F_{\theta}),d_1)$. A group $G \subseteq D, |G|>1$ is a cohesive-convergence group if there exists a value $k_0$ so that $P(A_{d_0, d_1} \cup B_{d_0, d_1}) =1, \forall d_0, d_1 \in G, K>k_0$. 

\subsection{Generative Group}
A cohesive convergence group $G$ is a generative group corresponding to $D_{train}$ if exists a member $d_0 \in G, d_0 \notin D_{train}$. 

\section{Observations}
To prepare for the experiments, the CIFAR-10 dataset will be divided into four parts as shown in Table \ref{tab1}. 
\begin{table}
\caption{Datasets}
\begin{center}\label{tab1}
\begin{tabular}{l@{\quad}ccr}
\hline
\multicolumn{1}{l}{\rule{0pt}{12pt}
                   Name}&\multicolumn{1}{c}{Symbol}&\multicolumn{1}{c}{Original}&\multicolumn{1}{r}{Size}\\[2pt]
\hline
Retain training set &  & training set & 49,488\\
Compact training set & A &  & 512\\
Retain test set & & test set & 9,488\\
Compact test set & B &  & 512\\[2pt]
\hline
\end{tabular}
\end{center}
\end{table}

Firstly, a neural network with the ResNet18 architecture \cite{he:kaiming} will be trained on the training set using the SGD optimization method \cite{ruder:sebastian, sutskever:ilya}. The training hyperparameters are as follows: the number of epochs is set to 32, batch size is 128, learning rate is 0.05, momentum is 0.9, and weight decay is 4e-3. 

\subsection{Existence of cohesive convergence groups}
Continue the experiment with the sampling step for two events A and B. Thus, the learning rate is adjusted to 0.001 and for each sampling step, the neural network will be undergone one training step with any batch of data from the training set. Then, apply the Algorithm \ref{alg:one} to obtain an evaluation table of the convergence-cohesion degree of any pair of elements belonging to the compact test set and the compact training set respectively. Finally, for each element in the compact test set, use the label of the element with the highest cohesive degree from the compact training set as the predicted label. The results of the experiment are presented in Table \ref{tab2}. 
\begin{table}
\caption{Accuracy of algorithms and corresponding target datasets.}
\begin{center}\label{tab2}
\begin{tabular}{l@{\quad}c|r}
\hline
Algorithm & Data set & Acuraccy\\[2pt]
\hline
1 & Compact test set & .93\\
2 & Compact test set & .75\\
arg max & training set & 1.\\
 & test set & .81\\[2pt]
\hline
\end{tabular}
\end{center}
\end{table}

The results show that the accuracy achieved by applying the algorithm is similar to the accuracy of applying $argmax$ on outputs of the neural network, called the argmax algorithm, over training samples. This demonstrates the existence of cohesive-convergence groups, where elements within the same group tend to share the same label. 

\subsection{Relationship of generative groups and bias-variance concept}
After the sampling step in Experiment 1, apply Algorithm \ref{alg:two} to obtain an evaluation table of test-side unconditional convergence-cohesion degree of any two elements from the compact training set and the compact test set, similar to Experiment 1. The reason of using the term `test-side unconditional convergence-cohesion degree' is because the ground truth labels is no longer used in calculating the loss values for elements in the compact test set, $L(F_{\theta},d \in B)$, as before. Instead, for each output value of the neural network corresponding to each predicted class, there will be a corresponding evaluation table of convergence-cohesion degree. Therefore, there will be 10 tables (or one table with one dimension more than the table in Experiment 1, this dimension has a size of 10). To achive the prediction label, for each element, return the label of the pair with the highest cohesive degree and has the identical corresponding labels. The result is presented in the second row of Table \ref{tab2}. 

Along with the previous results, the accuracies of algorithms using cohesive degree show a correspondence with the accuracies of the argmax algorithm for the compact training set and the compact test set. Because each prediction pairs include elements from the compact training set and the compact test set respectively, the results in this experiment also demonstrate the relationship between generative groups and the bias-variance concept of the neural network \cite{kohavi:ron:david,luxburg:ulrike:scholkopf}. 

\section{Algorithms}
\makeatletter
\newcommand\fs@norules{\def\@fs@cfont{\bfseries}\let\@fs@capt\floatc@ruled
  \def\@fs@pre{}%
  \def\@fs@post{}%
  \def\@fs@mid{\kern3pt}%
  \let\@fs@iftopcapt\iftrue}
\makeatother
\floatstyle{norules}
\restylefloat{algorithm}
\begin{algorithm}[H]
\caption{Sampling Cohesive-Degree Value algorithm}\label{alg:one}
\begin{algorithmic}[1]
\Require Neural network $N$, dataset $A$, dataset $B$
\Ensure $N=F_{\theta}$, $A$ is the compact training set, $B$ is the compact test set
\Procedure{SAMPLING\_BATCH}{$A$, $B$}\Comment{Retrieve a batch with elements mixed between set $A$ and set $B$, and the returned values are the separation of elements belonging to set A and elements belonging to set B along with their corresponding indices in $A$ and $B$}
\State Sampling $I_{A_0}, I_{A_0}[\cdot] \in [1..len(A)]$
\State Sampling $I_{B_0}, I_{B_0}[\cdot] \in [1..len(B)]$
\State $A_0 \gets A[I_{A_0}]$
\State $B_0 \gets B[I_{B_0}]$

\State \textbf{return} $A_0, B_0, I_{A_0}, I_{B_0}$
\EndProcedure

\Procedure{GET\_SCORE}{$L^A_0$, $L^A_1$, $L^B_0$, $L^B_1$}\Comment{Apply the sign function to the difference between $L^A_0$ and $L^A_1$, as well as $L^B_0$ and $L^B_1$. Then, multiply the results together (equivalent to the logical AND operator) to obtain the event $A \cup B$ for each element belonging to $A_0$, $B_0$. Accordingly, if the result is $1$, then the difference belongs to the event $A \cup B$; otherwise, it belongs to event $\overline{A \cup B}$}
\State $S^A \gets sign(L^A_0-L^A_1)$
\State $S^B \gets sign(L^B_0-L^B_1)$
\If {$dim(S^A) = dim(S^B)$}
	\State $S^{AB} \gets mul(S^A \times S^B)$
\Else
	\State $S^{AB} \gets []$
	\For {$i \in [0 .. 10]$}
		\State $S^{AB} \gets (S^{AB},mul(S^A[\cdot,\cdot,i] \times S^B))$
	\EndFor
\EndIf

\State \textbf{return} $S^{AB}$
\EndProcedure

\Procedure{SAMPLING}{$N$, $A$, $B$}
\State Initial a $len(A) \times len(B)$ 2D array $C$ with 0s
\For {$i \in [1..30]$}
\State Sampling $N_0 \gets T^{K}(N)$
\State Sampling $N_1 \gets T^{K+1}(N)$
\For {$j \in [1..len(A) \times len(B)]$}
\State $A_0, B_0, I_{A_0}, I_{B_0} \gets SAMPLING\_BATCH(A, B)$
\State $L^A_0 \gets L(N_0, A_0)$
\State $L^A_1 \gets L(N_1, A_0)$
\State $L^B_0 \gets L(N_0, B_0)$
\State $L^B_1 \gets L(N_1, B_0)$
\State $C[I_{A_0} \times I_{B_0}] \gets C[I_{A_0} \times I_{B_0}] + GET\_SCORE(L^A_0, L^A_1, L^B_0, L^B_1)$
\EndFor
\EndFor

\State \textbf{return} $C$
\EndProcedure
\end{algorithmic}
\end{algorithm}

\begin{algorithm}[H]
\caption{Sampling Test-side Unconditional-Cohesive-Degree Value algorithm}\label{alg:two}
\begin{algorithmic}[2]
\Procedure{SAMPLING}{$N$, $A$, $B$}
\State Initial a $len(A) \times len(B) \times 10$ 3D array $C$ with 0s
\For {$i \in [1..30]$}
\State Sampling $N_0 \gets T^{K}(N)$
\State Sampling $N_1 \gets T^{K+1}(N)$
\For {$j \in [1..len(A) \times len(B)]$}
\State $A_0, B_0, I_{A_0}, I_{B_0} \gets SAMPLING\_BATCH(A, B)$
\State $L^B_0, L^B_1 \gets L(N_0, B_0), L(N_1, B_0)$
\State $C[I_{A_0} \times I_{B_0}] \gets C[I_{A_0} \times I_{B_0}] + GET\_SCORE(N_0(A_0), N_1(A_0), L^B_0, L^B_1)$
\EndFor
\EndFor
\State \textbf{return} $C$
\EndProcedure
\end{algorithmic}
\end{algorithm}

\section{Conclusion}
This paper presents novel insights into the convergence dynamics of neural networks through the lens of cohesive-convergence groups. By delineating concepts, definitions, and algorithms, it sheds light on the intricate interplay between dataset structure and optimization outcomes. Experimental results validate the existence of cohesive-convergence groups, showcasing their utility in predictive tasks. Furthermore, the paper elucidates the relationship between generative groups and bias-variance concept, illuminating fundamental aspects of neural network behavior. These contributions advance the understanding of neural network convergence and pave the way for future research into more efficient and effective optimization strategies.

In addition to the conclusion generated by the ChatGPT chatbot above, aimed at increasing objectivity regarding the findings presented, an open-ended question that is: When generative groups imply that the convergence of one group may lead to the convergence of a larger group encompassing itself, which elements of the training set of CIFAR-10 would be included in the smallest cohesive-convergence group representing the entirety of the dataset? 


\end{document}